\documentclass{article}

     \usepackage[final]{neurips_2023}

\usepackage[utf8]{inputenc} 
\usepackage[T1]{fontenc}    
\usepackage{hyperref}       
\usepackage{url}            
\usepackage{booktabs}       
\usepackage{amsfonts}       
\usepackage{nicefrac}       
\usepackage{microtype}      
\usepackage{xcolor}         
\usepackage{enumitem}
\usepackage{tikz}
\usepackage{pgfplots}
\usepgfplotslibrary{statistics} 
\usepgfplotslibrary{fillbetween} 
\pgfplotsset{compat=newest} 
\usetikzlibrary {arrows.meta}
\usepackage{tikzscale}
\usepackage{enumitem}

\newcommand{\Figref}[1]{Fig.~\ref{#1}}
\newcommand{\Tabref}[1]{Tab.~\ref{#1}}
\newcommand{\Secref}[1]{Sec.~\ref{#1}}

\setcitestyle{square}

\title{Exploring Dataset-Scale Indicators of Data Quality}

\author{%
  Benjamin Feuer \\
  New York University\\
  \texttt{bf996@nyu.edu} \\
  \And
  Chinmay Hegde \\
  New York University\\
  \texttt{ch3773@nyu.edu} \\
}

\begin{document}

\maketitle

\begin{abstract}
    Modern computer vision foundation models are trained on massive amounts of data, incurring large economic and environmental costs. Recent research has suggested that improving data quality can significantly reduce the need for data quantity. But what constitutes data quality in computer vision? We posit that the quality of a given dataset can be decomposed into distinct sample-level and dataset-level constituents, and that the former have been more extensively studied than the latter. We ablate the effects of two important dataset-level constituents: label set design, and class balance. By monitoring these constituents using key indicators we provide, researchers and practitioners can better anticipate model performance, measured in terms of its accuracy and robustness to distribution shifts.
\end{abstract}


\section{Introduction}

\subsection{Motivation} 

In recent years, the field of computer vision has seen remarkable progress in a range of sub-disciplines~\citep{li_semantic-sam_2023, mildenhall_nerf_2021, rombach2021highresolution}. Much of this progress has derived from \textit{foundation models}~\citep{yuan_florence_2021, Radford2021LearningTV}. Modern computer vision foundation models are trained on massive amounts of data, and the general trend has been to achieved improved performance by scaling up dataset size.

However, increasing the size of the dataset is not the only way to improve the downstream performance of a given model. Recent work has introduced into the literature the importance of \textit{data quality}, a phrase which is meant to encompass all facets of a dataset which impact downstream performance, aside from its size~\citep{qualitynotquantity, isacaption, gadre_datacomp_2023}.

The question of \emph{which} facets of data quality are relevant for improved performance, however, is under-explored in the literature. Even less frequently studied is whether there exist \emph{quantifiable, predictive indicators} of these facets. If reliable indicators of dataset quality can be found prior to training, then researchers can better estimate the impact of modifications to their datasets (or even guide dataset design), and ultimately lead to reduction of human, environmental, and economic costs of large model training ~\citep{sharir_cost_2020,changpinyo2021cc12m,bommasani_opportunities_2022}.

\subsection{Our Contributions} 
In this (preliminary) work, we present several new findings on the constituents of dataset quality. Unlike most previous works, our aim is to discover indicators that are properties at the dataset-scale, rather than measures of image-level quality. Via a series of controlled ablation studies, we explore the downstream performance impact of certain constituents, and show that these can be used as predictive indicators of the quality of datasets prior to model training. In brief:

\begin{enumerate}[noitemsep, parsep=0pt, left=0pt]
    \item We conduct thorough ablation studies on two important dataset properties --- \emph{label set size}, and \emph{class imbalance} --- and analyze the (sometimes) complex effects they have on downstream metrics in image classification tasks.
    \item For each of the above two constituents, we provide a list of \textit{key indicators} which are predictive of model performance. Our indicators are inexpensive to compute, scalable to very large dataset sizes, and can be identified prior to model training.
\end{enumerate}

Our results can be viewed as building blocks towards a systematic taxonomy of the notion of ``quality'' at the dataset scale, which may enable improved design choices for datasets in computer vision. 

\section{Related Work}
\label{sec:related-work}

\paragraph{Dataset curation.}The question of how best to curate a dataset from a raw, large set of images is an important problem in computer vision. The common approach is to use {full supervision}; classes are chosen in advance, raw samples are algorithmically/manually filtered, and then manually labeled by annotators, and (sometimes) class balance is enforced; the classic example is ImageNet \citep{Imagenet}. Such datasets tend to produce very strong baselines \citep{kornblith_better_2019}. Other approaches take a more relaxed approach to filtration, assigning unfiltered web-scraped images to human labelers and applying a wide range of possible class labels~\citep{Kuznetsova2018TheOI}. 

As datasets scale from millions to billions of samples, human labeling becomes impractical. While early approaches such as \citet{Thomee2016YFCC100MTN} initially curated datasets without any supervision, \textit{weak supervision} strategies have now become popular. With no human in the loop, proxy measures of quality become essential. Sample-level quality indicators include encoding format, size, aspect ratio, and offensive content~\citep{sharma-etal-2018-conceptual}. Labeling strategies for these images sometimes rely on image tags from large social image-sharing sites \citep{sun_revisiting_2017, LimitsOfWeakly}.

One challenge in studying dataset curation in computer vision derives from the fact that many of the largest image datasets cited in the literature are not publicly avaialable \citep{sun_revisiting_2017, LimitsOfWeakly, qualitynotquantity} A notable recent exception to this is the work of \citet{LAION400m}. Subsequent works such as \citep{openclip, gadre_datacomp_2023, feuer_distributionally_2023} have taken advantage of these developments to train new models and design data-centric challenges.

\paragraph{Indicators of data quality.}There have been many proposed approaches to predict the behavior of models for a given test set, coalescing in a NeurIPS competition in December 2020 \citep{pmlr-v133-jiang21a}. Broadly, submitted methods fell into three meta-categories: (i) generalization measures derived from theoretically motivated generalization bounds; (ii) data augmentation methods, which estimate the generalization error of a trained model by computing its accuracy on synthetic data, and (iii) geometry and statistics of intermediate representations. The most successful approach was that of  \citep{natekar2020representation}, which used a combination of (ii) and (iii). Unlike these works, our method is agnostic both to the choice of model and to the contents of individual samples (assuming a strong correspondence between images and their labels).

\paragraph{Distributional robustness.}An important dimension in which foundation models have been found to outperform smaller computer vision models --- and which we use as a key metric in our experiments --- is \emph{distributional robustness}, a test-time paradigm which aims to estimate model robustness to distribution shifts \citep{Recht2019DoIC}. A \emph{distribution shift} is defined as evaluation data which differs from the data on which a model was trained due to natural factors. Real world image classifiers require predictable model behavior under such shifts. Models trained on large, heterogeneous datasets tend to provide greater distributional robustness than their counterparts trained on less data \citep{feuer_distributionally_2023}.

\paragraph{Hard negatives.}Another related but distinct line of inquiry is the use of hard negatives, which have been shown to positively impact accuracy and robustness in recommender systems and, more recently, contrastive learners. In recommender systems which rely on implicit feedback, only positive classes are directly observed. The remaining data is a mixture of actually negative and missing values, which can then be \textit{mined} for valuable instances for learning.~\citep{hard-negs-shi, rendle2012bpr} Unlike these works, we investigate the more common non-contrastive, fully supervised learning setting, in which all classes are presumed to be known at training time.

\paragraph{Class imbalance.}In real-world data collection scenarios, some events are common while others are exceedingly rare, which commonly results in class-imbalanced datasets. It has long been observed that such datasets underperform compared to balanced ones~\citep{GoodBengCour16}, and  comprehensive meta-analyses have demonstrated that oversampling, the dominant method for remedying class imbalance, is ineffective on neural networks.~\citep{class-imbal-buda-2018} Despite this long-standing awareness, both the exact source of this problem and its resolution remain unclear.~\citep{shwartz-ziv2023on} Unlike prior works, ours attempts to produce quantitative measures of class imbalance which will be predictive of downstream performance, allowing for more targeted use of what remedies are avaiable.

\section{Experimental Setup}
\label{sec:experimental-design}

\textbf{Measurable indicators of dataset quality. }As discussed above in \Secref{sec:related-work}, most existing works which attempt to evaluate data quality do so at the \emph{sample level}; they focus on properties pertaining to individual samples, such as image resolution and image fidelity. Our focus instead is on exploring holistic \emph{dataset-level} properties, which are typically determined by the dataset's creators and (if that dataset is used as a benchmark) are typically treated as immutable. 

In this short paper, we focus on understanding the predictive power of two such properties: (i) number of classes (label set size), and (ii) number of samples per class (individual class size). We chose these properties since they are ubiquitous, concretely measurable, and tend to have large impacts on the performance of the corresponding trained models.

\textbf{Data scaling. }We define \textit{horizontal scaling} (H-scaling) as scaling up the label set size in the dataset while holding individual class size constant. We define \textit{vertical scaling} (V-scaling) as scaling up individual class size while holding the label set size constant. 

\textbf{Architecture. }Our baseline architecture against which all variations are compared is a modified ResNet-50 with a 1000-class linear classification head~\citep{ResNet}. The specifics of the modifications are described in \citep{openclip}. Whenever we use an architecture other than our standard baseline, we refer to it by its name in the timm library from \citet{rw2019timm}.

\textbf{Pretraining datasets. } The experiments described in this paper were primarily conducted using the JANuS dataset, introduced by \citep{feuer_distributionally_2023}. JANuS is a composite image-caption dataset sourced from four different datasets of origin. Every sample in JANuS contains one or more labels (for training conventional computer vision models) and one or more captions (for training vision-language models) with varying degrees of supervision. These properties make it an ideal dataset for conducting controlled experiments for H-scaling and V-scaling. In addition to JANuS, we make use of LAION, ImageNet and OpenImages, from \citet{Imagenet, Kuznetsova2018TheOI, LAION400m}, respectively; further details are provided in the experiments below.

\textbf{Training details.} In our experiments, we train with mixed precision, at a batch size of 256, and do not use gradient clipping. We use the AMP library to implement the training process. Learning rate is chosen once, via grid search, for each new architecture / dataset pair. Models are typically distributed across a single node with 4 NVIDIA A100 GPUs. All models are trained for 256 epochs. Following ~\citet{isacaption}, we use SimCLR augmentations (resize, crop, flip, jitter, blur, grayscale) rather than CLIP augmentations (resize and crop) for model training. 
A few of our models are not trained from scratch, but are instead evaluated zero-shot using weights sourced from \citet{rw2019timm}; we note this whenever it is the case.

\textbf{Labels.} We evaluate on a single, broad-scope label set of 100 classes corresponding to ImageNet-100 (IN100), which is the first constituent dataset of JANuS. In our tables, we refer to the validation set for IN100 as IN100-Val, and the average shift accuracy as IN100-Avg. Rob. OI100 refers to OpenImages-100, the second constitutent dataset of JANuS.

\noindent\textbf{Distribution Shifts.} Folliwing the literature, whenever we measure robustness, we report the average of four shifts on ImageNet. \textit{ImageNet-V2} was designed to duplicate, as closely as possible, the original ImageNet test set~\citep{Recht2019DoIC}. \textit{Imagenet-Sketch} is a distribution shift covering sketches, paintings, drawings and illustrations~\citep{Wang2019LearningRG}. \textit{Imagenet-R} is a 200-class subset of ImageNet-2012 focused on renditions of everyday objects~\citep{hendrycks2021many}. \textit{Imagenet-A} is a 200-class subset of ImageNet-2012 which was algorithmically selected~\citep{hendrycks2021nae}.

\textbf{Data filtration. }We define \emph{data filtration} as any strategy which sub-selects from a larger pool of possible samples. A simple example of a filtration strategy would be conducting a web search for the target classes in a dataset, and selecting the first $k$ samples in the search.

\textbf{Metrics for distributional robustness. }
Our primary metric is \emph{average robustness} (abbv: Avg. Rob.), which is the average test-set accuracy of a model on a set of distribution shifts. Although this measure is easy to interpret, it can conceal substantial performance differences between shifts. 

\section{Results}

The fact that both label set size and class (im)balance impact image classification models should be folklore to computer vision practitioners. However, to the best of our knowledge, these two properties have not been directly contrasted, given an overall data budget. We address the following questions:

\begin{enumerate}[nosep,left=0pt]
    \item \textbf{For a given overall data budget, is it better to scale up individual class sizes (V-scaling), or to scale the number of classes (H-scaling)?} See \Secref{sec:label-set}.
    \item \textbf{How does class imbalance impact accuracy and robustness?} See \Secref{sec:class-bal}.    
\end{enumerate}

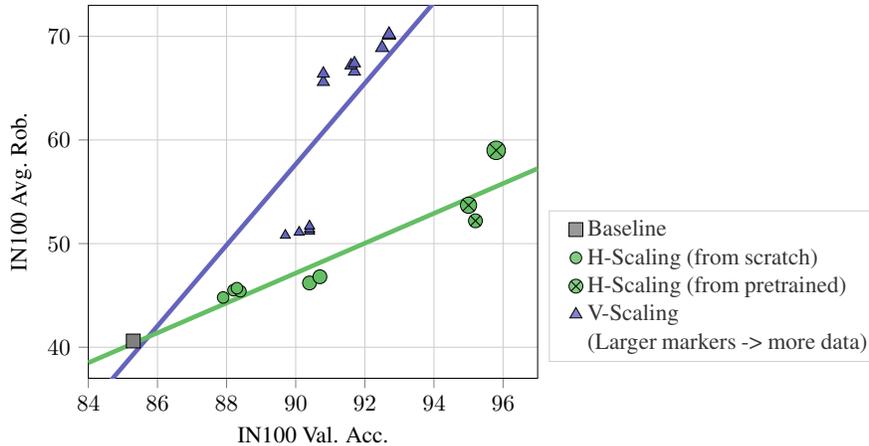
\begin{figure}[!t]
    {
    \centering
    \resizebox{.85\linewidth}{!}{

\begin{tikzpicture}

\definecolor{darkgray176}{RGB}{176,176,176}
\definecolor{darkred}{RGB}{139,0,0}
\definecolor{green}{RGB}{0,128,0}
\definecolor{blue204}{RGB}{200,200,250}
\definecolor{lightgray204}{RGB}{204,204,204}
\definecolor{blue128}{RGB}{100,100,190}
\definecolor{lightgray128}{RGB}{128,128,128}
\definecolor{blue90}{RGB}{60,60,160}
\definecolor{green204}{RGB}{200,250,200}
\definecolor{green170}{RGB}{160,220,160}
\definecolor{green128}{RGB}{100,190,100}
\definecolor{green90}{RGB}{60,160,60}
\definecolor{green60}{RGB}{45,120,45}
\definecolor{green40}{RGB}{30,90,30}

\def\discon{$\wr\!\!\!\wr$}

\begin{axis}[
legend cell align={left},
legend style={
  fill opacity=0.8,
  draw opacity=1,
  text opacity=1,
  at={(1.025,0.45)},
  anchor=north west,
  draw=lightgray204
},
xmin=84, xmax=97,
ymin=37, ymax=73,
tick align=outside,
tick pos=left,
x grid style={lightgray204},
ylabel={IN100 Avg. Rob.},
xmajorgrids,
y grid style={lightgray204},
xlabel={IN100 Val. Acc.},
ymajorgrids,
]

\addplot [name path=upper, domain=84:97, samples=2, blue128, forget plot, draw=none] {3.904*x + -295.7};

\addplot [domain=84:97, samples=2, blue128, line width=2, forget plot] {3.904*x + -293.7}; 

\addplot [name path=lower, domain=84:97, samples=2, blue128, forget plot, draw=none] {3.904*x + -291.7};

\addplot [domain=84:97, samples=2, green128, line width=2, forget plot] {1.442*x + -82.63}; 


\addplot [only marks, mark=square*, mark size=3.0, draw=black, mark options={fill=lightgray128}]
table{
    x y
    85.3 40.6
};
\addlegendentry{Baseline}



\addplot [only marks, mark=*, mark size=2.5, draw=black, mark options={fill=green128}]
table{
    x y
    87.9 44.8
    88.2 45.5
    88.4 45.4
    88.3 45.7
};
\addlegendentry{H-Scaling (from scratch)}

\addplot [only marks, mark=*, mark size=3.0, draw=black, mark options={fill=green128}, forget plot]
table{
    x y
    90.4 46.2
    
    90.7 46.8 
};

\addplot [only marks, mark=otimes*, mark size=3.0, draw=black, mark options={fill=green128}]
table{
    x y
    95.2 52.2 
};
\addlegendentry{H-Scaling (from pretrained)}

\addplot [only marks, mark=otimes*, mark size=3.5, draw=black, mark options={fill=green128}, forget plot]
table{
    x y
    95.0 53.7
};

\addplot [only marks, mark=otimes*, mark size=4.0, draw=black, mark options={fill=green128}, forget plot]
table{
    x y
    95.8 59.0
};


\addplot [only marks, mark=triangle*, mark size=2.5, draw=black, mark options={fill=blue128}, forget plot]
table{
    x y
    89.7 50.8
    90.1 51.1
    90.4 51.2
    90.4 51.4
    90.4 51.7
};



\addplot [only marks, mark=triangle*, mark size=3.0, draw=black, mark options={fill=blue128}]
table{
    x y
    90.8 65.6
    90.8 66.4
    91.7 66.6
    91.6 67.2
    91.7 67.4
};
\addlegendentry{V-Scaling}


\addplot [only marks, mark=triangle*, mark size=3.25, draw=black, mark options={fill=blue128}, forget plot]
table{
    x y
    92.5 68.9
    92.7 70.1
    92.7 70.2
};

\addlegendimage{empty legend}
\addlegendentry{(Larger markers -> more data)}


\end{axis}

\end{tikzpicture}
    }
    \caption{\sl\textbf{Models benefit from scaling data horizontally, as well as vertically.}  It is well known that vertical scaling (increasing the number of samples for in-distribution classes) improves accuracy (x-axis) and robustness (y-axis). Surprisingly, we find that horizontal scaling, increasing the number of out-of-distribution samples, also helps with robustness, and improves accuracy \textit{more} than vertical scaling, even when H-scaling classes are drawn from a different distribution than target classes -- see also \Tabref{tab:scaling-ablations-ds}. Dot size represents size of training dataset (vertical scaling is ID, horizontal is OOD). Darker points represent models with more parameters. Image best viewed in color.}
    \label{fig:label-set-size-001}
    }
\end{figure}

\subsection{Label Set Size}
\label{sec:label-set}

Scaling up dataset size has become the \emph{de facto} driving force for improving the accuracy (and robustness) of image classification models. But what is the \emph{right} way to scale up datasets: should we just scale up samples per class? Or are there benefits if the model is trained on a larger set of classes? To fairly compare these two choices, we keep the test label set constant. The net effect of the latter case is that the model sees image examples that are OOD with respect to the test set, and we zero out the logits of the OOD classes at test time.

We enumerate results on two distinct sets of large scale classification models. All vertically scaled models are trained by us. For horizontal scaling, we employ a mix of pretrained and from-scratch (with the larger models generally evaluated from pretrained checkpoints). We distinguish between them in \Figref{fig:label-set-size-001} with distinct labels. 

\begin{table}[b]
    \centering
    \resizebox{\textwidth}{!}{\begin{tabular}{c c c c c}
    \toprule
    \textbf{Model} & \textbf{Dataset Size} & \textbf{Training Stages} & \textbf{IN100-Val / Avg. Rob.} \\
    \midrule
    \textbf{resnetv2\_50x1\_bit} & 15.4 Mn & PT 21800 FT 1000 ZS 100 & 95.8\% / 59.0\% \\
    \textbf{resnetv2\_50} & 12.6 Mn & PT 1000 FT 12000 ZS 100 & 95.0\% / 53.7\% \\
    \textbf{resnetv2\_50} & 1.2 Mn & PT 1000 ZS 100 & 95.2\% / 52.2\% \\
    \textbf{swin\_base\_patch4\_window7} & 14.2 Mn & PT 21800 ZS 100 & 86.7\% / 43.4\% \\
    \textbf{flexivit\_base\_1000ep} & 14.2 Mn & PT 21800 ZS 100 & 68.9\% / 18.8\% \\
    \textbf{convnext\_base} & 14.2 Mn & PT 21800 ZS 100 & 66.4\% / 36.2\% \\
    \textbf{resnetv2\_50x1\_bit} & 14.2 Mn & PT 21800 ZS 100 & 26.7\% / 5.2\% \\
    \bottomrule \\
    \end{tabular}}
    \caption{\sl \textbf{Horizontal scaling with large label sets.} We find that horizontal scaling with large label sets works much better when training occurs in multiple stages. Surprisingly, the process works almost as well whether the larger label set comes first or second; label set size is not an inherent obstacle to the success of horizontal scaling. Model names correspond to those found in the timm library~\citep{rw2019timm}. PT (pretraining) refers to the first stage of model training. FT (fine-tuning) refers to subsequent stages of model training. ZS (zero-shot) refers to the evaluation process described in \Secref{sec:label-set-size-findings}. Parameter count is rounded to the nearest million; label set size to the nearest hundred.}
    \label{tab:scaling-ablations}
\end{table} 

\subsubsection{Key Indicators}
A natural indicator for vertical scaling is per-class dataset size (though this approach demands separate attention to class balance; see \Secref{sec:class-bal}). Training label set size is the indicator for horizontal scaling.

\subsubsection{Findings}
\label{sec:label-set-size-findings}

\emph{Surprisingly, we find that models achieve more robust and accurate overall representations via horizontal scaling.} It is well known that models benefit from seeing more ID samples during training, and that contrastive models require positive as well as large batch negative samples to learn~\citep{bendavid_uml, Radford2021LearningTV}. Less well-studied is whether non-contrastive models also benefit from scaling batch negatives (OOD classes, in this case). We find that increasing the number of OOD classes from 0 to 900 leads to reliable gains in both accuracy and robustness. In fact, models gain \textit{more} ID accuracy from horizontal scaling (OOD) than vertical scaling (ID).

\begin{table}[h]
    \centering
    \resizebox{0.6\textwidth}{!}{\begin{tabular}{c c c}
    \toprule
    \textbf{OOD Data Source} & \textbf{Dataset Size} & \textbf{IN100-Val / Avg. Rob.} \\
    \midrule
    \textbf{OpenImages} & 1.2 Mn & 91.5\% / 50.2\% \\
    \textbf{ImageNet} & 1.3 Mn & 90.7\% / 46.8\% \\
    \textbf{FractalDB} & 1 Mn & 85.4\% / 40.2\% \\
    \bottomrule \\
    \end{tabular}}
    \caption{\sl \textbf{Horizontal scaling with different out-of-distribution approaches.}  Horizontal scaling works as well or better when the out-of-distribution classes are not from ImageNet. Horizontal scaling on synthetic out-of-distribution classes underperforms scaling on natural images. Dataset size is rounded to the nearest 100,000.}
    \label{tab:scaling-ablations-ds}
\end{table} 

\emph{When we pretrain and fine-tune models, even very large label sets can benefit from horizontal scaling.} As the label set grows extremely large, our naive approach to horizontal scaling fails; we examine a range of timm model checkpoints pretrained on the entirety of ImageNet (which contains almost 22,000 classes), even larger and better-performing architectures fail to match much smaller models fine-tuned on IN1000  (see \Tabref{tab:scaling-ablations}). 

There are many potential explanations for this phenomenon. Candidates include the lack of a validation set against which to optimize, or the extreme class imbalance. Another possible explanation is that horizontal scaling cannot succeed when the class space is very large. We show that this is not the case -- multi-stage training, even on very large label sets, (see \Tabref{tab:scaling-ablations}, lines 1, 2)  outperforms naive horizontal scaling on 1000 classes (\Tabref{tab:scaling-ablations}, line 3).

The fact that pretraining and fine-tuning seems to work in \textit{either direction} is particularly surprising to us. Starting with fewer classes and fine-tuning on many more works almost as well as the more intuitive approach of starting with a model pretrained on many classes and fine-tuning on a much smaller number of classes. We leave further exploration of these dynamics to future work.

\emph{Horizontal scaling works with non-ImageNet images.} In (\Tabref{tab:scaling-ablations-ds}), we describe the results of horizontal scaling on both ImageNet and non-ImageNet out of distribution classes. Specifically, we experiment with adding classes from OpenImages which do not overlap with ImageNet-1000 classes, and synthetic classes from the FractalDB dataset~\citep{KataokaACCV2020}. We find that the OpenImages classes actually \textit{out-perform} the ImageNet classes substantially, despite the dataset being slightly smaller. This finding suggests that dataset blending, combined with horizontal scaling, is a promising approach for training more performant computer vision models.

\subsection{Class (Im)Balance}
\label{sec:class-bal}

In line with previous results (see  \Secref{sec:related-work}), we confirm that class imbalance has a substantial negative effect on both validation accuracy and average robustness for IN100 classification. We extend this discussion by exploring a distinct, but related question -- \textit{why} do imbalanced datasets underperform? 

Inspired by recent work investigating Zipfian distributions~\cite{chan_data_2022}, we posit two hypotheses. Zipf's law stipulates that the frequency of an event is inversely proportional to its rank in a frequency table. Zipfian distributions occur in natural language, where a small number of words (like “the” and “and”) occur very frequently, while the majority of words occur rarely. We interrogate each of these hypotheses: first, that the underperformance is due to the existence of a few overrepresented classes, a property which we call \textbf{left-skewedness}; second, that the underperformance is due to the existence of long-tail classes with very few samples in them, which we call \textbf{long-tailedness}.

\subsubsection{Key indicators}

We propose two potential key indicators to better compare these hypotheses. Our proposed indicator for left-skewedness is the percentage of samples in the dataset which are members of the most common k\% of classes. For the problems described in this paper, we heuristically set $k=5$. In a perfectly balanced dataset, then, left-skewedness will be 5\%. In OI100 without rebalancing, it is 64\%.

Our proposed indicator of long-tailedness is the percentage of classes with fewer than $k$ samples in them  (\%<$k$ Classes). We heuristically explore two choices: $k = 500$ and $k = 100$.

\subsubsection{Findings}

\begin{table}[h]
    \centering
    \resizebox{\textwidth}{!}{\begin{tabular}{c c c c c}
    \toprule
    \textbf{Data Source} & \textbf{Dataset Size} & \textbf{Left-skew} & \textbf{Long-tail @ 500 / Long-tail @ 100} & \textbf{IN100-Val / Avg. Rob.} \\
    \midrule
    \textbf{in100 (100\%)} & 125,000 & 5\% & 0\% / 0\% & 85.3\% / 40.6\% \\
    \textbf{in100 (62\%)} & 130,000 & 5\% & 0\% / 0\% & 82.5\% / 43.1\% \\
    \textbf{oi100 (71\%)} & 190,000 & 45\% & 0\% / 0\% & 82.2\% / 44.3\% \\
    \textbf{oi100 (60\%)} & 101,000 & 13\% & 0\% / 0\% & 79.3\% / 41.3\% \\
    \textbf{oi100 (88\%)} & 90,000 & 25\% & 0\% / 0\% & 76.6\% / 38.8\% \\
    \textbf{oi100 (67\%)} & 135,000 & 31\% & 9\% / 9\% & 73.9\% / 40.7\% \\
    \textbf{oi100 (57\%)} & 105,000 & 12\% & 9\% / 9\% & 73.4\% / 39.1\% \\
    \textbf{oi100 (100\%)} & 135,000 & 64\% & 64\% / 9\% & 67.7\% / 37.2\% \\
    \textbf{oi100 (100\%)} & 53,000 & 18\% & 67\% / 9\% & 58.2\% / 31.1\% \\ 
    \bottomrule \\
    \end{tabular}}
    \caption{\textbf{Under-represented classes trigger performance declines.} \sl We ablate class imbalance by blending samples from ImageNet and OpenImages, which share a data source. Surprisingly, class imbalance alone does not cause model performance to degrade (Left-skew, the percentage of samples which are in one of the 5 most common classes, is not predictive of performance). Rather, it is the existence of long-tail classes containing very few samples (Long-tail @ k). The 9 longest tailed classes in OI100 ($k=100$) account for the majority of performance decline. Sample sizes are rounded to the nearest 1,000 samples. Percentiles are rounded to the nearest percentile.}
    \label{tab:attrib-classbal}
\end{table} 

Our main results on class imbalance can be found in \Tabref{tab:attrib-classbal}, where we report data source, dataset size, our key indicators, validation accuracy, and average accuracy under shift.

\emph{Surprisingly, we find that class imbalance alone does not trigger a performance decline.} When we preserve the imbalance in the largest classes of OpenImages, but rebalance small classes, performance \textit{improves} slightly (see \Tabref{tab:attrib-classbal} line 3) compared to the baseline (\Tabref{tab:attrib-classbal} line 8). Furthermore, when we decrease the degree of imbalance in OpenImages by truncating the largest classes, performance degrades (see \Tabref{tab:attrib-classbal} line 9). We conclude that models improve when training on more samples of a given class, even when classes are imbalanced.

\emph{Rather, the performance declines are attributable the presence of underrepresented classes.} In \Tabref{tab:attrib-classbal}, our long-tailedness indicator is in perfect rank-order agreement with IN100 validation accuracy. Dataset size and left-skewedness exhibit only weak agreement.

\emph{Furthermore, the very smallest classes account for most of the decline.} We ablate this by rebalancing all classes with more than $k=100$ samples (using this approach, 91 of 100 classes are balanced). We find that so doing accounts for only 39\% of the overall accuracy gains. The majority of accuracy loss from class imbalance is attributable to the very small classes. 

\section{Conclusions and Future Work}

In this paper, we outline useful indicators of data quality which are inexpensive to compute and can be used during pretraining. 
    In \Secref{sec:label-set}, we showed that horizontal scaling benefits models trained from scratch. Based on this, for small datasets, as an alternative to pretraining and fine-tuning, we suggest co-training on target classes and classes drawn from existing image datasets in order to ensure that the overall label set size is large.
    In \Secref{sec:class-bal}, we showed that underrepresented classes, rather than class imbalance, harms model performance. In a setting with limited resources, we recommend rebalancing the classes with the fewest samples first.

In future work, we hope to introduce additional important dataset-level factors that influence pretraining and provide indicators which make use of the image space as well as the caption space.

{
\bibliographystyle{neurips_2023}
\bibliography{neurips_2023}
}

\end{document}